Large Language Models' Varying Accuracy in Recognizing Risk-Promoting and Health-Supporting Sentiments in Public Health Discourse: The Cases of HPV Vaccination and Heated Tobacco Products

Soojong Kim[1], Kwanho Kim[2], and Hye Min Kim[3]

[1] Department of Communication, University of California Davis, United States

[2] Department of Media, College of Politics and Economics, Kyung Hee University, South Korea

[3] Department of Communication, University of Massachusetts Boston, United States

**Author Note**

Soojong Kim  https://orcid.org/0000-0002-1334-5310

Correspondence concerning this article should be addressed to Soojong Kim, 1 Shields Ave, Kerr Hall #361, Davis, CA 95616, United States. Email: sjokim@ucdavis.edu

**Abstract**

Machine learning methods are increasingly applied to analyze health-related public discourse based on large-scale data, but questions remain regarding their ability to accurately detect different types of health sentiments. Especially, Large Language Models (LLMs) have gained attention as a powerful technology, yet their accuracy and feasibility in capturing different opinions and perspectives on health issues are largely unexplored. Thus, this research examines how accurate the three prominent LLMs (GPT, Gemini, and LLAMA) are in detecting risk-promoting versus health-supporting sentiments across two critical public health topics: Human Papillomavirus (HPV) vaccination and heated tobacco products (HTPs). Drawing on data from Facebook and Twitter, we curated multiple sets of messages supporting or opposing recommended health behaviors, supplemented with human annotations as the gold standard for sentiment classification. The findings indicate that all three LLMs generally demonstrate substantial accuracy in classifying risk-promoting and health-supporting sentiments, although notable discrepancies emerge by platform, health issue, and model type. Specifically, models often show higher accuracy for risk-promoting sentiment on Facebook, whereas health-supporting messages on Twitter are more accurately detected. An additional analysis also shows the challenges LLMs face in reliably detecting neutral messages. These results highlight the importance of carefully selecting and validating language models for public health analyses, particularly given potential biases in training data that may lead LLMs to overestimate or underestimate the prevalence of certain perspectives.

Large Language Models' Varying Accuracy in Recognizing Risk-Promoting and Health-Supporting Sentiments in Public Health Discourse: The Cases of HPV Vaccination and Heated Tobacco Products

The advancement of Artificial Intelligence (AI) is transforming numerous fields, including medical sciences, public health, and health communication. AI technologies are especially promising for understanding public perceptions of critical and emerging health issues. Especially, Large Language Models (LLMs) have demonstrated remarkable capabilities in conducting human-level decision-making and data processing, making them suitable for interpreting complex text data (Katz et al., 2023; Liu et al., 2023) and opening new possibilities for understanding public perceptions on health issues more effectively.

Yet, little is known about how LLMs perform when tasked with analyzing public perceptions across different health topics and communication formats. LLMs, trained on vast amounts of text data, often reflect and amplify existing patterns in public discourse where potentially biased and harmful information receives disproportionate attention from the general public and media outlets (Bai et al., 2025; M. H. J. Lee et al., 2024; Ling et al., 2025). Claims that pose greater societal risks often garner heightened attention and interest, leading to more extensive documentation, discussion, and analysis (e.g., Edwards, 2022; Hauck, 2019; Tingley, 2021), and this can make such information more distinctly recognizable to LLMs. Additionally, health messages opposing promoted health behaviors and practices often employ more salient linguistic patterns and emotional appeals, which could lead to systematic differences in how accurately LLMs detect and classify different health viewpoints (Faasse et al., 2016; Kearney et al., 2019; J. Lee et al., 2024). Thus, similar to AI systems exhibiting inconsistent performance across demographic groups due to representation imbalances in training data (Buolamwini and Gebru, 2018; Koenecke et al., 2020), LLMs may show disparate accuracy in detecting content and claims that can pose risks to public health versus those aligned with public health guidelines, which is a concern that directly motivated the present research.

Understanding these potential biases is crucial for researchers who use advanced computational tools — automated techniques such as machine learning classifiers and LLMs, which can be used to process and classify digital data efficiently with minimal human intervention — for objective analysis of public health discourse. Because the extent to which such biases appear in LLM outputs can vary depending on the model's architecture, training data, and calibration methods, it is also important for researchers to compare different LLMs when conducting public health analyses. While many models reflect patterns in public discourse that may lead to biased outcomes, not all LLMs exhibit the same tendencies, and developers are increasingly implementing strategies to mitigate these limitations.

The current research aims to systematically examine these potential discrepancies by evaluating how LLMs perform in detecting different types of health-related sentiments. Specifically, we focused on two distinct sentiment categories: risk-promoting and health-supporting. In the present research, we defined risk-promoting sentiments as sentiments that support, justify, or encourage actions or beliefs associated with increased risk in a public health context, such as anti-vaccination sentiment, and support for tobacco products and their use. These sentiments oppose preventive health measures, encourage potentially harmful behaviors, and often involve spreading inaccurate or misleading information (e.g., Hornik et al., 2022; Jolley & Douglas, 2014). In contrast, health-support sentiments are those that support, justify, or encourage actions or beliefs associated with reduced public health risks, such as pro-vaccination views and rejection of tobacco product use. These sentiments typically align with public health guidelines and recommendations promoted by health professionals and public health agencies.

To evaluate how LLMs perform in detecting these two types of health-related sentiments, this research examines three major LLMs—OpenAI's GPT, Google's Gemini, and Meta's LLAMA—in analyzing sentiments related to Human Papillomavirus (HPV) and heated tobacco products (HTPs), across two major social media platforms, Twitter and Facebook. These topics were chosen because they represent distinct public health challenges where both sentiments opposing and supporting promoted health behaviors play crucial roles in shaping public discourse, making them ideal cases for investigating potential asymmetries in LLMs' sentiment detection capabilities.

To ensure the relevance of the comparison, we chose three LLMs that represent the current landscape of widely adopted language models. GPT and Gemini are the two most widely used proprietary LLMs in the United States. According to a recent survey, 72% of users reported using GPT, and 50% reported using Gemini, reflecting their broad adoption across both personal and professional contexts (Rainie, 2025). LLAMA is the only open-source model among the top five most used LLMs in the U.S. (Rainie, 2025), with 20% of users reporting its use. Its open-access nature has enabled wide adoption, customization, and integration into a range of applications, making it especially valuable for research and development. LLAMA is also recognized as the most performant open-source LLM (Meta AI, 2024). Including these three models thus enables a robust and balanced evaluation of leading LLMs across both proprietary and open-source paradigms, enhancing the relevance and applicability of findings across diverse technological and user contexts.

**Examining Public Perceptions of Health Issues Using Computational Methods**

Evaluating public perceptions of health issues is crucial in public health and health communication research. By examining sentiments, emotions, and topics of public discourse on specific health issues, scholars and professionals can monitor trends in health behaviors and perceptions in communities and society, which helps inform the design of health communication campaigns and interventions as well as assess their effectiveness (Hornik et al., 2022; Kearney et al., 2019; Kwon and Park, 2020). The analysis of public perceptions also aids in adjusting strategies to improve public awareness, engagement, and acceptance of specific health practices and policies (Himelboim et al., 2020; Lakon et al., 2016; Poirier and Cobb, 2012). Furthermore, it enables the identification of areas of resistance or concern, guiding the development of more tailored resources and approaches (Budhwani et al., 2021; Jamison et al., 2019).

Potential discrepancies in accuracy in identifying risk-promoting sentiments versus health-supporting ones could lead to biased results in the analysis of public perception of health issues that are critical, emerging, and socially controversial. For instance, if a computational tool detects risk-promoting messages at a higher level of accuracy than health-supporting ones, the analysis may overestimate the prevalence of risk-promoting public sentiments on a health issue. This skewed representation could lead to misguided efforts to design and implement health campaigns and interventions aimed at improving public acceptance and compliance with health measures.

The present research investigates how accurately LLMs identify risk-promoting and health-supporting sentiments, focusing on two public discourses about critical health issues: Human Papillomavirus (HPV) vaccination and heated tobacco products (HTPs). By examining LLMs across the two important health topics, we aim to assess their effectiveness more broadly, understanding potential biases and discrepancies in LLM-based sentiment analysis. The complexity and importance of public perceptions of these contentious health topics, explained in the following, also underscore the need for precise and balanced analysis.

**The Cases of HPV Vaccination and HTPs**

Vaccination continues to be a polarizing subject, eliciting a broad spectrum of opinions ranging from strong endorsement of its health benefits to persistent skepticism and resistance (Blane et al., 2023; Sturgis et al., 2021; Yaqub et al., 2014). Since these diverse perspectives are linked to individual health behaviors and policy support, studying public opinions on vaccination is crucial for social scientists and public health experts (Cruickshank et al., 2021; Paul et al., 2021; Yaqub et al., 2014). HPV vaccination, in particular, continues to be a subject of public discussion and debate. Although HPV vaccination is recognized as a vital preventive measure against various cancers and its benefits are well-established (Shing et al., 2022), it still faces strong public opposition and skepticism (Dunn et al., 2017; Sonawane et al., 2021). Understanding public sentiment toward HPV vaccination is thus crucial for shaping effective public campaigns and health interventions. A past study also showed that variations in different sentiments toward HPV vaccination were significant enough to track trends over time, based on the analysis of 1.4 million social media posts related to HPV vaccination (Du et al., 2020).

HTPs, which heat processed tobacco leaves to deliver nicotine-containing aerosol, are gaining global market share. Unlike e-cigarettes, which vaporize liquid nicotine, HTPs use real tobacco that is heated rather than combusted (Centers for Disease Control and Prevention, 2024). This emerging tobacco product prompts considerable interest from researchers and policymakers due to its implications for tobacco prevention and cessation efforts (Abroms et al., 2024). While HTPs are often marketed as less harmful alternatives to combustible tobacco, government authorities, tobacco prevention scientists, and public health practitioners strongly discourage their use overall, upholding this claim only in very limited contexts, primarily when assisting current tobacco users in quitting (American Academy of Pediatrics et al., 2024; Centers for Disease Control and Prevention, 2024; Department of Health and Social Care, 2022).

Social media platforms, where stakeholders frequently discuss tobacco regulation through policy updates, advertisements, and consumer feedback, are known to serve as critical arenas for the discussions of HTPs (Abroms et al., 2024; Jun et al., 2022; O'Brien et al., 2020). A number of previous studies analyzed HTP-related messages collected from social media and other digital platforms, with a particular focus on marketing and promotion messages and HTP regulation-related discussions, to identify sentiments and message topics of the public discourse (Abroms et al., 2024; Barker et al., 2021; Berg et al., 2021; Gu et al., 2022; Jun et al., 2022; O'Brien et al., 2020).

Based on this, the present research considered anti-vaccination sentiment in the context of HPV vaccination and pro-HTP sentiment in the HTP context as risk-promoting sentiments. On the other hand, pro-vaccination sentiment for HPV vaccination and anti-HTP sentiment for HTPs were considered health-supporting sentiments.

**Computational Approaches to Analyzing Large-Scale Health Discourse**

The immense scale of text data generated from various digital platforms provides rich resources for analyzing public perceptions. Systematic analyses of these large datasets have become critical for researchers and policymakers seeking a comprehensive and accurate understanding of public attitudes toward health issues. However, while large-scale text provides valuable information, the sheer volume makes manual analysis impractical, necessitating the use of computational methods.

Scholars have made considerable methodological advancements in processing and drawing inferences from large-scale text datasets. Prior to the rise of LLMs, natural language processing techniques based on machine learning classifiers were widely used to filter and cluster large datasets (Huang et al., 2014; Shapiro et al., 2017; Valdez et al., 2023). Indeed, a review study found that, among 755 studies analyzing online health information between 2005 and 2020, a substantial number utilized computational methods, with 31% applying machine-learning techniques (Shakeri et al., 2021). For example, using machine learning, a past study analyzed over 80 million messages about tobacco products and categorized them by themes and sentiment (Kim et al., 2020). More recent evidence suggests that machine-learning tools outperform other traditional methods, such as dictionary-based techniques, in analyzing health-related messages (van Atteveldt et al., 2021). Machine-learning tools continue to improve with the development of new algorithms, including pre-trained models (Baker et al., 2022; Khan et al., 2023; Rustam et al., 2021).

**Past Approaches Based on Machine-Learning Classifiers.** Computational approaches have become essential for analyzing large-scale health-related discourse, as the increasing scale of datasets makes manual analyses less feasible. Recent studies on HPV vaccination, for instance, used machine-learning methods to classify social media messages into different categories. Du et al. (2020) used a deep-learning classifier trained on 6,000 manually annotated tweets out of 1.4 million. Using the classifier, the researchers identified that positive sentiment toward the HPV vaccine steadily increased while negative sentiment gradually declined. Boucher et al. (2023) trained a machine-learning model to explore how vaccine-hesitant and vaccine-confident online communities discuss HPV vaccination. They found more negative tweets among the hesitant community, and the proportion of both sentiments increased in both communities following the WHO's declaration of the COVID-19 Public Health Emergency.

Studies on tobacco products also demonstrate the utility of automated sentiment analysis methods. A meta-analysis of 74 tobacco prevention studies found that 43% used machine-learning techniques (Fu et al., 2017). However, most of these efforts focused on combustible cigarettes and e-cigarettes (e.g., K. Kim et al., 2020a), and the research on recently introduced tobacco products rapidly gaining popularity, such as HTPs, has largely relied on manual analysis of relatively small datasets (e.g., Jun et al., 2022; M. Kim et al., 2024). This suggests the potential utility of more advanced computational approaches, such as LLMs.

**Potentials of LLMs as a Methodological Breakthrough.** The emergence of LLMs, such as OpenAI's GPT, has opened new possibilities. Trained on vast amounts of text, these models have demonstrated capabilities for human-level reasoning and decision-making (Brown et al., 2020; Katz et al., 2023; Liu et al., 2023; Thoppilan et al., 2022). With improvements in usability and accessibility, LLMs are increasingly used in public health and health communication research. Prior approaches often relied on machine-learning classifiers, which enabled large-scale analysis of public discourse far beyond the reach of manual content analysis. However, these methods typically required substantial investment in developing task-specific classifiers and preparing annotated reference datasets. In contrast, LLMs offer the potential to conduct similar analyses with little or no task-specific training, thereby reducing both time and resource demands (S. Kim et al., 2024; Ziems et al., 2023). Recent studies have demonstrated the feasibility of LLMs for analyzing diverse health-related data, including medical records, academic publications, and patient-generated content, highlighting their potential for automated analyses of public discourse on health issues based on digital data (Chen et al., 2024; Clusmann et al., 2023; Park et al., 2024).

### Research Aim and Research Questions

The main aim of the present research is to evaluate LLMs' accuracy in detecting different sentiments in public discourses on health issues on social media. By doing so, this research investigates the following four research questions.

1. How accurately do LLMs classify sentiments compared with human evaluators, across health issues, platforms, and models? We evaluate the accuracy of LLMs, which indicates the degree of alignment between LLM and human evaluations of sentiments in messages, across different health issues, language models, and digital platforms.

2. Do LLMs vary in accuracy when detecting risk-promoting versus health-supporting sentiments? We investigate whether there is a discrepancy in the accuracy of detecting risk-promoting and health-supporting sentiments on health issues. This examination will help identify possible biases or limitations of LLMs in recognizing different perspectives, which could hinder balanced interpretations of public perceptions.

3. How does accuracy vary depending on the specific health issue, language model, or platform? This comparison will reveal how these factors might affect public perception analysis using LLMs, highlighting the importance of carefully selecting AI tools and considering their strengths and limitations in particular contexts. Understanding these differences also helps identify which language models are best suited for specific issues and communication styles.

4. How well do LLMs detect neutral sentiment, relative to clearly supportive or opposing sentiments? Neutral sentiment, which includes balanced information presenting both supportive and opposing perspectives on a health issue, often makes up a considerable portion of public discourse (Liu and Liu, 2021) and presents unique challenges in sentiment analysis, compared to more direct, strong sentiments, such as anti-vaccination or pro-HTP viewpoints (Koppel and Schler, 2006).

It is important to note that estimating the relative prevalence of sentiments was not the primary objective of this research. The main aim of this study was to evaluate the accuracy of the newly emerging AI technique in detecting different sentiments, and all features of the present research were carefully determined to achieve that aim. For example, we compared the same number of opposing and supporting sentiment messages across health issues and message formats, which allowed us to ensure that the evaluation of the tool's performance was balanced and not affected by an uneven distribution of sentiment types. Although we anticipate that the current findings will ultimately contribute to understanding the prevalence of different sentiments related to specific health issues, this research was focused solely on assessing the performance of the tool itself.

## Methods

### Data Collection

Social media messages related to the two distinct public health issues were identified from two social media platforms supporting different message formats: Facebook, which supports long-form content, and Twitter (now renamed "X"), which supports short-form content. These social media messages were identified and verified in previous studies (Kim and Kim, 2025; S. Kim et al., 2024). Specifically, for the data collection of Facebook posts, we used CrowdTangle, a social media analytics tool owned by Facebook's parent company, Meta. CrowdTangle provided researchers with access to a comprehensive historical database of over 7 million public Facebook Pages and Groups, making it a valuable resource for conducting large-scale searches and data retrieval (CrowdTangle, 2023). To collect Twitter messages, we used Twitter's API version 2.0 (Twitter, 2023). Specifically, we used their academic research access program that allowed researchers to access the entire historical archive of all Twitter messages. It is important to note that since the completion of the present research, access to CrowdTangle has been discontinued, and Twitter's academic API has become materially more restricted, limiting access to historical social media data for research purposes.

Regarding HPV vaccination, we identified Facebook posts that included keywords related to HPV vaccination — ((("hpv" OR "papillomavirus" OR "cervical") AND ("vaccine" OR "vaccination" OR "vax" OR "shot" OR "jab")) OR "gardasil" OR "cervarix" — and were published during a 10-year period from January 2012 to December 2021. Using the same search criteria and timeframe, we also identified Twitter messages. These searches resulted in 141,479 posts from Facebook and 676,193 from Twitter.

For HTPs, we conducted a keyword search to collect Facebook posts containing ("heat not burn" OR "heat-not-burn" OR "heated tobacco" OR "tobacco heating" OR "iqos"). In addition, we identified posts containing ("HTP" OR "HNB") but only when these posts also contain one of the following words: "smoking", "smoke", "vaping", "vape", "tobacco", "cig", and "nicotine." This search targeted posts published during an 8-year period from January 2014 (the year that the first HTP product was launched) through December 2021. The same search criteria and period were used to identify messages on Twitter. These searches resulted in 16,284 Facebook posts and 60,031 tweets.

The Institutional Review Board of the University of California Davis granted an exemption for this research under application 2031428-1.

**Human-Evaluated Message Sets**

From these four message pools, we curated a total of 1,600 messages with human-evaluated sentiment labels, comprising 400 messages for each health issue and platform, equally distributed between risk-promoting ($n = 200$; i.e., anti-vaccination and pro-HTPs) and health-supporting ($n = 200$; i.e., pro-vaccination and anti-HTPs) sentiments. The procedure of message selection and human evaluation followed methodologies established in previous studies (Kim and Kim, 2025; S. Kim et al., 2024), and detailed information is provided in Supplementary Online Material (SOM). We call these four sets of human-verified messages "message sets." Each message was reviewed by three human experts and assigned the corresponding label. Drawing an equal number of messages across the two sentiment categories and the four message pools allowed a balanced evaluation of LLM performance that was not affected by an uneven distribution of sentiment types.

Following the analysis of health-supporting and risk-promoting sentiments, we conducted a post hoc analysis to assess the accuracy of large language models in detecting more ambivalent and subtle sentiments. For this additional analysis, we curated an additional 400 messages labeled as "neutral" by human evaluators, comprising 100 neutral messages per health issue and platform.

**LLM Evaluation**

For each message in a message set, we created a prompt and directed an LLM to classify the message into one of five categories (anti, pro, neutral, mixed, and irrelevant), identical to those used in human evaluation. Each prompt included instructions, the content of the message, and an explanation of the categories (Refer to SOM for the prompts used for this research). We generated a total of 96,000 LLM responses for this research, calculated as follows: 3 language models × 2 health topics × 2 platforms × 2 sentiment categories × 200 messages per category ×

VARYING ACCURACY OF LLM IN RECOGNIZING HEALTH SENTIMENTS                1520 LLM response instances per message. For the post hoc analysis of neutral sentiment, we generated additional 24,000 LLM responses: 3 language models × 2 health topics × 2 platforms × 100 messages × 20 LLM response instances per message. We used the API for each language model rather than personal chat sessions through web interfaces, and each LLM response was generated using a newly initiated, stateless task to ensure that no prior prompts or outputs influenced subsequent LLM responses.

The full methodological procedure is illustrated in Figure S1 in the SOM. Example risk-promoting and health-supporting messages are provided in Table S5 in the SOM.

*Accuracy Calculation*

We conducted the AI evaluation of each message based on 20 LLM responses regarding its sentiment. LLMs generate responses by predicting the next word in a sequence based on the preceding context, using probabilistic sampling over a distribution of likely tokens derived from the model's learned parameters (Brown et al., 2020; Jurafsky and Martin, 2009). The inherent randomness in the sampling process can lead LLMs to produce varying responses to the same prompt. This variability can be beneficial for analyzing human-generated content: By utilizing multiple AI-generated responses, researchers can explore a range of potential interpretations of a message and make a more informed decision based on these diverse responses, similar to having multiple human coders analyze the same content (Heseltine and Clemm Von Hohenberg, 2024; S. Kim et al., 2024).

In this research, we produced 20 LLM response instances for each message, and this set of response instances was considered as a pool of potential machine evaluations for the message. A possible machine evaluation of a message's sentiment was determined by randomly selecting three among the 20 response instances and identifying the majority response within the three. In the event of a tie (for example, selecting one each of anti-HTPs, pro-HTPs, and neutral response instances), an additional instance was randomly chosen until a decisive majority was established. Considering the random nature of the subset selection process, the selection and majority determination process was simulated 1,000 times for each message. Each iteration involved recording whether the machine's decision aligned with the human evaluation of the message (recorded as 1) or not (recorded as 0). This record was then averaged across all iterations to compute the mean accuracy, which ranges between 0 and 1 and serves as a measure of the language model's precision in classifying a specific message. The present research did not define a specific threshold for acceptable accuracy, as the focus was on comparative performance across conditions rather than absolute benchmarks. However, it should be noted that very high values, such as those exceeding .8, could reasonably be viewed as indicating very good accuracy in sentiment classification tasks.

In the present research, the standard error of the mean accuracy is not reported because it becomes arbitrarily small as the number of simulations increases. The ability to generate an extremely large number of simulations would result in a narrow confidence interval, but this would not reflect the true uncertainty in the LLM's accuracy in real-world sentiment classification. Thus, reporting standard errors or confidence intervals would not meaningfully capture the LLM's actual classification error in predicting sentiments.

***Language Models***

We assessed the performance of three LLMs: GPT-4 Turbo, version 2024-04-09 (OpenAI, 2024), Gemini-1.0 Pro (Google Gemini Team, 2024), and LLAMA-3 (Meta AI, 2024). The examination of multiple language models aimed to compare patterns across different LLMs and elucidate their individual strengths and weaknesses. These models stood out not only for their capabilities (Vellum, 2024) but also for their accessibility and recognition, being products of leading tech giants—Google, Meta, and OpenAI (backed by Microsoft). Each model utilizes a unique architecture and was trained on distinct datasets. Despite these differences, scholars across social sciences and public health have acknowledged the significance of these models as powerful tools for social data analysis (Argyle et al., 2023; S. Kim et al., 2024; Ziems et al., 2023).

## Results

### Overall Accuracy of LLMs

The results showed considerable capabilities of LLMs in detecting both risk-promoting and health-supporting sentiments across different health issues and message formats. Figure 1 presents the accuracy of the language models in detecting sentiments in Facebook messages, and Figure 2 presents the accuracy in Twitter messages. Each grey dot in these figures represents a value of accuracy calculated from a single iteration of the simulation. With the exception that Gemini's accuracy for health-supporting sentiment fell below .8 for both Facebook (.688) and Twitter (.704), the accuracy of detecting HPV vaccination sentiments was above .8 across different sentiments and models, ranging between .825 (GPT for risk-promoting sentiment on Twitter) and .980 (LLAMA for risk-promoting sentiment on Facebook). On the other hand, the overall accuracy for detecting HTP-related sentiments was generally lower than HPV's, yet remained above .7 and reached as high as .925 (LLAMA for health-supporting on Facebook), with one exception: Gemini's accuracy for risk-promoting sentiment on Twitter was .692.

### Accuracy by Sentiment Type and Platform

The results reveal systematic discrepancies in the accuracy of detecting risk-promoting and health-supporting sentiments. First, LLMs showed higher accuracy in identifying risk-promoting sentiment in both HPV vaccination and HTPs for Facebook messages. Specifically, for HPV vaccination messages on Facebook (Figure 1-A), higher accuracy than health-promoting sentiments in detecting risk-promoting sentiment was observed for all three LLMs tested: GPT (risk: .928, health: .837), Gemini (risk: .917, health: .684), and LLAMA (risk: .980, health: .916). Regarding Facebook HTP messages (Figure 1-B), GPT (risk: .799, health: .780) and Gemini (risk: .841, health: .723) demonstrated consistently higher accuracy for risk-promoting sentiment, while LLAMA (risk: .892, health: .925) showed higher accuracy for the health-promoting sentiment.

Regarding Twitter messages, LLMs showed higher accuracy in identifying health-supporting sentiment across HPV vaccination and HTPs. Specifically, for HPV vaccination messages on Twitter (Figure 2-A), higher accuracy in detecting health-supporting sentiment was observed for GPT (risk: .825, health: .904) and LLAMA (risk: .896, health: .951), while Gemini (risk: .862, health: .705) showed higher accuracy for risk-promoting sentiment. For HTP vaccination messages on Twitter (Figure 2-B), higher accuracy in detecting health-supporting sentiment was observed for all three LLMs tested: GPT (risk: .727, health: .739), Gemini (risk: .692, health: .719), and LLAMA (risk: .829, health: .839).

**Model Comparisons**

It is worth noting that LLAMA consistently demonstrated higher accuracy for both risk-promoting and health-supporting sentiments compared with GPT and Gemini. This performance advantage was especially evident for HPV vaccination messages on Facebook, where LLAMA achieved an accuracy of .980 for risk-promoting sentiment and .916 for health-supporting sentiment, and for HTP messages on Facebook, in which LLAMA recorded .892 and .925, respectively. A similar pattern emerged on Twitter, where LLAMA outperformed GPT and Gemini for both health-supporting sentiment (HPV vaccination: .951; HTP: .839) and risk-promoting sentiment (HPV vaccination: .896; HTP: .829).

**Performance on Neutral Sentiment**

Additional analysis on neutral sentiment revealed challenges for LLM-based classification when detecting more subtle and ambivalent sentiments. Overall, the accuracy of detecting neutral sentiment was consistently lower than that for risk-promoting and health-supporting sentiments across all cases tested in this study. For HPV vaccination messages, all three models showed noticeably weaker performance in identifying neutral sentiment than in detecting risk-promoting or health-supporting sentiments: LLAMA (Facebook: .311, Twitter: .392), GPT (Facebook: .488, Twitter: .438), and Gemini (Facebook: .662, Twitter: .672). This trend persisted for HTP messages: LLAMA (Facebook: .330, Twitter: .418), GPT (Facebook: .709, Twitter: .786), and Gemini (Facebook: .687, Twitter: .567). Figures S5 and S6 in SOM present these results alongside the accuracy rates for health-supporting and risk-promoting sentiments reported earlier.

Figure 3 visualizes the differences between human and LLM evaluations, focusing on GPT's and LLAMA's performance for HTP messages (Specific values are presented in Table 1.) It shows how the same messages were classified by human evaluators and the language model. The widths of the lines connecting identical sentiments on the left and right sides represent the number of messages where human and machine classifications align. The results show that most misclassifications occurred when human-classified risk-promoting (orange on the left) or health-supporting messages (blue on the left) were misclassified as neutral by the language model (green on the right). This is evident from the substantial connections between risk-promoting and neutral, as well as health-supporting and neutral. As discussed earlier, these misclassifications and the resulting reduction in accuracy were more pronounced in Twitter messages than in Facebook messages, as reflected in the thicker lines linking risk-promoting and neutral, as well as health-supporting and neutral. LLAMA's superior performance compared with GPT is also explained by fewer misclassifications into neutral, as indicated by narrower connections between risk-promoting/health-supporting and neutral for LLAMA than GPT. Comparisons for other cases, available in SOM, also indicate that misclassifications primarily resulted from the language models incorrectly categorizing messages as neutral.

## Discussion

The findings of the present research offer several important insights into the use of LLMs for analyzing sentiments in online public health discourses, and its potential strengths and weaknesses. Overall, LLMs can achieve considerable accuracy in detecting risk-promoting and health-supporting sentiments across health issues and digital platforms. Although all three LLMs tested in this study showed promising results, variations in their performance highlight the need for researchers to consider specific model characteristics and contextual factors when applying these AI tools. First, all three LLMs demonstrated considerable capabilities in detecting different sentiments in online public health discourses. Across both HPV vaccination and HTP topics, and on both Facebook and Twitter, the majority of accuracy scores for risk-promoting and health-supporting sentiments surpassed a .7 level, with several instances exceeding .8. For example, LLAMA achieved an accuracy of .980 for risk-promoting HPV vaccination messages on Facebook, representing one of the highest rates recorded in the current research. These findings highlight LLMs' feasibility for large-scale analyses of public health issues. From a practical standpoint, these results offer promising evidence for scholars and practitioners seeking to leverage automated sentiment analysis for monitoring and studying public health discourse.

Second, the current research reveals systematic patterns of discrepancies in the accuracy of detecting risk-promoting versus health-supporting sentiments. Specifically, for Facebook messages, most models demonstrated higher accuracy in detecting risk-promoting sentiments than health-supporting ones. Contrarily, for Twitter messages, higher accuracy often emerged for health-supporting sentiments.

Observed discrepancies in accuracy may reflect, in part, differences in linguistic patterns and context richness across message formats and platforms, but further empirical research is needed to establish the specific mechanisms underlying these gaps. These findings suggest that LLM performance may vary across social media environments, highlighting the importance of validating model behavior in relation to platform structure. These findings align with earlier research on algorithmic biases, which suggests that uneven representation of particular content types across platforms can affect how AI models detect and classify inputs (Buolamwini and Gebru, 2018; Koenecke et al., 2020). Our results thus caution that LLM-based sentiment analyses could overestimate or underestimate the prevalence of certain health perspectives if models are not carefully validated or adjusted.

Third, we found notable variations across health issues, media platforms, and language models, consistent with emerging work on LLM applications in the health domain (S. Kim et al., 2024; Ziems et al., 2023). Overall, models tended to show higher accuracy for HPV vaccination messages compared with HTP messages. The discrepancy could be attributable to more extensive public discussion and dataset availability on vaccination topics, making training data for such content more abundant. HTP-related discussions, which are relatively newer and possibly more niche, may involve linguistic patterns or terminologies less frequently encountered by the models during training. The results also suggest that platform-specific factors shaped classification accuracy. Facebook messages, which allow for longer-form writing, yielded higher accuracy for detecting risk-promoting sentiments, compared with short-form Twitter content. Longer messages may contain additional context and semantic cues, facilitating more reliable classification. In other words, although all three LLMs tested in this study showed promising results, variations in their performance highlight the need for researchers to consider specific model characteristics and contextual factors when applying these AI tools. Furthermore, among the three tested LLMs, LLAMA consistently achieved the highest or near-highest accuracy rates, especially for Facebook messages on HPV vaccination. GPT demonstrated robust performance but still fell behind LLAMA in all cases tested in this study. Gemini generally performed well but exhibited notable dips in health-supporting sentiment detection. These distinctions highlight the importance of model choice in applied health communication research.

In an additional analysis, we observed lower accuracy scores for neutral sentiment detection in all tested cases, sometimes falling below .50. This trend was identified for both HPV vaccination and HTP content, on Facebook and Twitter, and for all three models. The difficulties in classifying neutrality mirror broader literature on sentiment analysis, where the absence of clear polar expressions can make neutral messages more difficult to detect. The nuanced or context-dependent language in many neutral posts may reduce the models' accuracy. Methodologically, researchers need to take into account such systematic weaknesses, especially if neutral expressions form a substantive portion of health discourse. Potential solutions include expanding training datasets with content explicitly labeled as neutral, refining language model prompts, or integrating human oversight in challenging cases.

Recent advances, such as Retrieval-Augmented Generation (RAG) architectures and the integration of knowledge graphs, are actively being explored to enhance the precision of LLM classification by enriching model inputs and outputs with external knowledge sources, often based on methods such as fine-tuning or the incorporation of external modules (Lewis et al., 2020; Yang et al., 2024). Although these innovations were beyond the scope of the current research, we acknowledge that such developments represent promising directions for future work aimed at mitigating classification errors and biases.

The current results highlight the practical importance of tailoring model choice and evaluation strategies to a specific communication context. For instance, researchers analyzing short-form content such as tweets may benefit from complementing LLM-based classification with additional validation steps or hybrid approaches that combine automated and manual coding. Future research should systematically test how structural features of messages interact with model architectures to influence classification outcomes.

Beyond methodological considerations, this research also has broader implications for the role of LLMs in domains traditionally shaped by qualitative research. Their capacity to process and classify large-scale text data introduces new opportunities to analyze public discourse with efficiency and consistency. This shift calls for rethinking how qualitative insights are generated, balancing computational scale with human-centered interpretation. Crucially, the present findings that LLMs vary in their accuracy across sentiment types, platforms, and health issues call attention to the need to treat these tools not as replacements, but as complements to qualitative inquiry. Recent studies have shown that while LLMs offer substantial promise in automating aspects of qualitative analysis, they also raise concerns about accuracy, context sensitivity, and the potential erosion of researcher agency (Barros et al., 2025; Roberts et al., 2024). When carefully validated and used, LLMs can support and extend traditional approaches, offering a scalable foundation for deeper interpretive analysis in public health research.

VARYING ACCURACY OF LLM IN RECOGNIZING HEALTH SENTIMENTS                     24The current research is not without limitations. First, while focusing on two health issues and two platforms provides insights across multiple contexts, it may still limit generalizability to other health topics or social media environments, such as Instagram, TikTok, or YouTube. Second, although we intentionally selected balanced samples of risk-promoting and health-supporting messages to assess model accuracy without distributional biases, real-world data commonly exhibit uneven sentiment distributions. Future research may also evaluate LLM performance in more naturally imbalanced datasets. Third, even though we employed multiple replications of each classification through a majority mechanism, further comparative studies could test alternative approaches to reduce variability. Lastly, since the completion of this research, access to CrowdTangle has been discontinued, and the academic API for Twitter has become considerably more limited. These restrictions pose challenges for future research aiming to replicate or expand the current work and highlight the need for alternative data access strategies or collaborations with platform providers. Future studies should also aim to create benchmark datasets of online messages to facilitate the evaluation and comparison of various AI-based methods in the public health context, while addressing the privacy, legal, and ethical challenges associated with releasing social media content.

The current research demonstrates that LLMs, despite their overall strong performance in classifying risk-promoting and health-supporting content, display meaningful variation across issues, platforms, and model types. Although LLMs show potential as efficient tools for large-scale sentiment analysis, persistent discrepancies remain. Understanding and addressing these challenges is critical in advancing health communication, public health, and policy research, enhancing their accuracy, efficiency, and timeliness in this rapidly evolving era of AI.

**Supplementary Material**

Supplementary Material is available online (https://osf.io/pcxsz/).

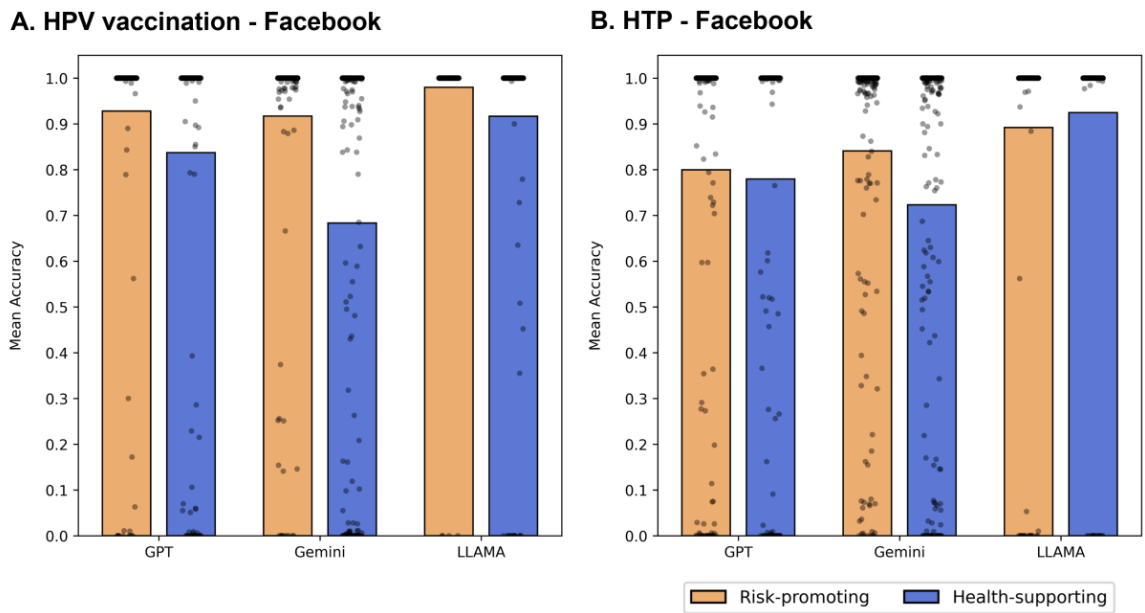

Figure 1. LLM Accuracy for Facebook Messages by Health Issue and LLM Model

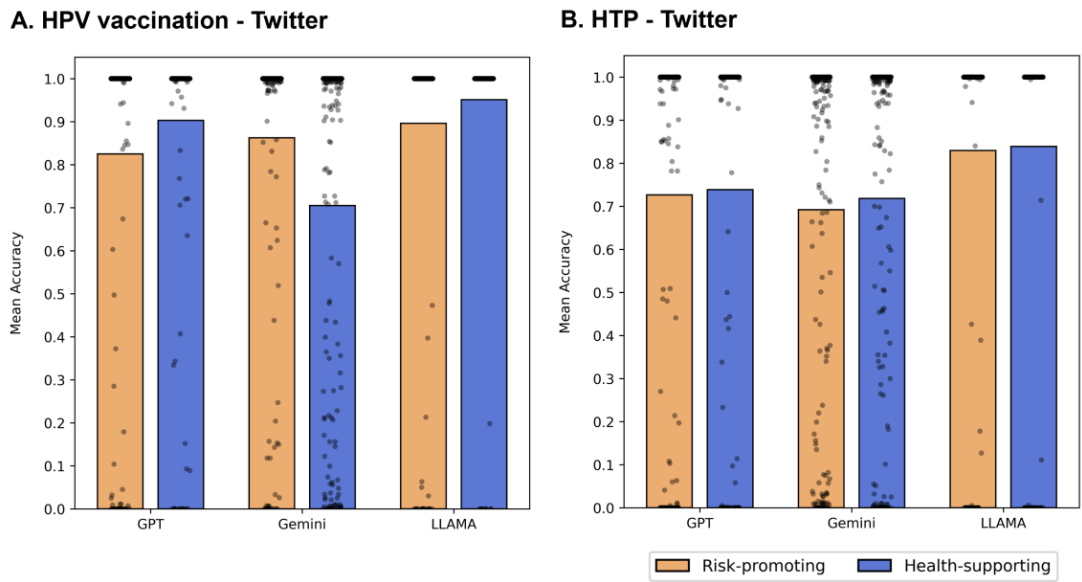

Figure 2. LLM Accuracy for Twitter Messages by Health Issue and LLM Model

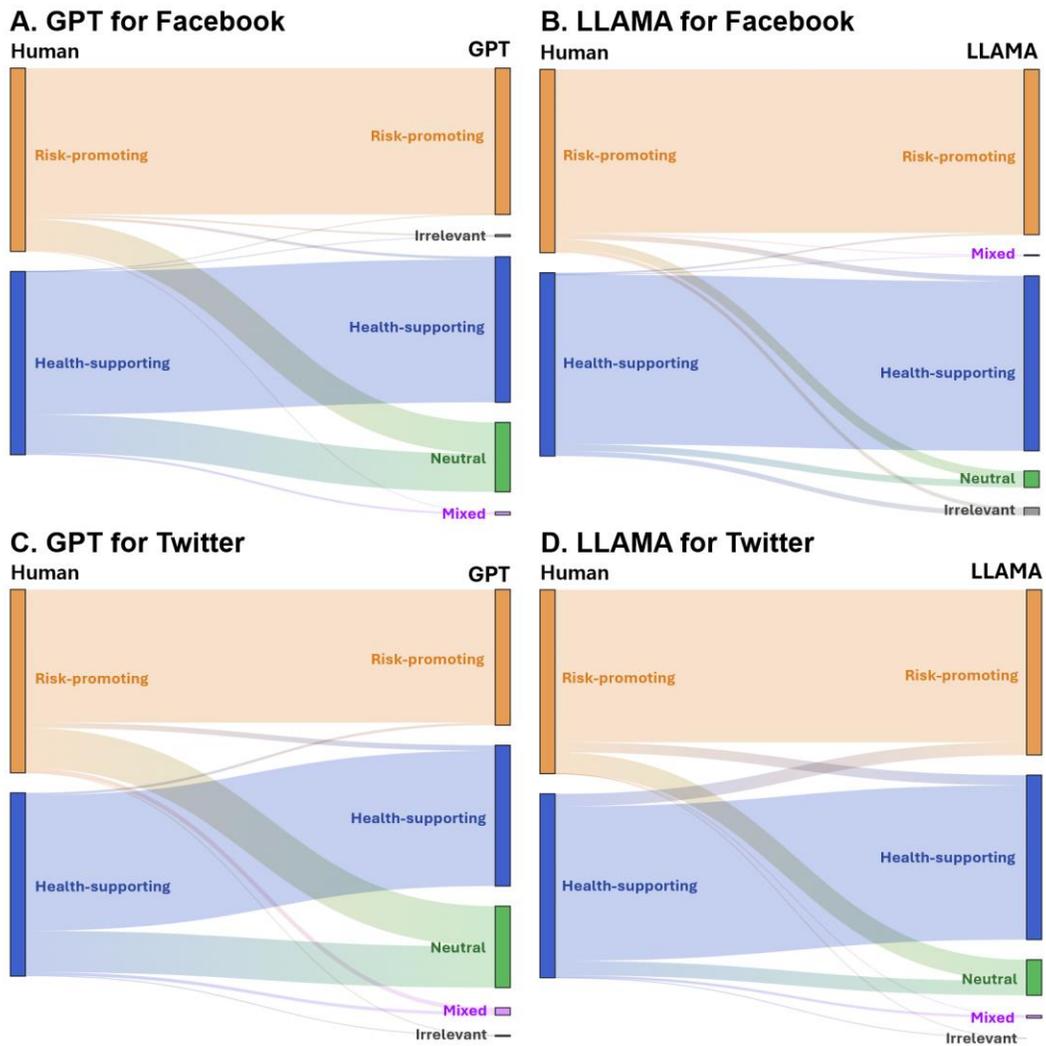

Figure 3. Comparison of Human and LLM Classification Results for HTP Vaccination Messages

VARYING ACCURACY OF LLM IN RECOGNIZING HEALTH SENTIMENTS        34Table 1. Comparison of Human and Machine Evaluations for HTP Messages: GPT and LLAMA

| GPT with Facebook messages | | | LLAMA with Facebook messages | | |
|---|---|---|---|---|---|
| GPT | Human | | LLAMA | Human | |
| | Risk | Health | | Risk | Health |
| Risk | 159.8 (79.9%) | 0.007 (0.004%) | Risk | 178.4 (89.2%) | 2.0 (1.0%) |
| Health | 3.0 (1.5%) | 155.9 (78.0%) | Health | 6.0 (3.0%) | 184.9 (92.5%) |
| Neutral | 34.2 (17.1%) | 41.6 (20.8%) | Neutral | 10.5 (5.2%) | 7.6 (3.8%) |
| Mixed | 1.0 (0.5%) | 2.3 (1.2%) | Mixed | 1.0 (0.5%) | 0.0 (0.0%) |
| Irrelevant | 2.0 (1.0%) | 0.1 (0.1%) | Irrelevant | 4.1 (2.0%) | 5.5 (2.7%) |
| Total | 200 (100%) | 200 (100%) | Total | 200 (100%) | 200 (100%) |
| GPT with Twitter messages | | | LLAMA with Twitter messages | | |
| GPT | Human | | LLAMA | Human | |
| | Risk | Health | | Risk | Health |
| Risk | 145.2 (72.6%) | 2.8 (1.4%) | Risk | 165.9 (82.9%) | 13.8 (6.9%) |
| Health | 5.9 (3.0%) | 147.7 (73.9%) | Health | 11.0 (5.5%) | 167.8 (83.9%) |
| Neutral | 43.9 (22.0%) | 45.0 (22.5%) | Neutral | 23.0 (11.5%) | 15.7 (7.8%) |
| Mixed | 4.6 (2.3%) | 3.4 (1.7%) | Mixed | 0.1 (0.1%) | 2.7 (1.4%) |
| Irrelevant | 0.3 (0.1%) | 1.0 (0.5%) | Irrelevant | 0.0 (0.0%) | 0.0 (0.0%) |
| Total | 200 (100%) | 200 (100%) | Total | 200 (100%) | 200 (100%) |

*Note*. The table includes mean counts and column proportions. "Risk" refers to risk-promoting, and "Health" refers to health-supporting.